\documentclass[10pt,twocolumn,letterpaper]{article}

\usepackage{wacv}
\usepackage{times}
\usepackage{epsfig}
\usepackage{graphicx}
\usepackage{amsmath}
\usepackage{amssymb}

\usepackage{bm}
\usepackage{xcolor}
\usepackage{soul}
\usepackage{multirow}
\usepackage{tabulary}
\usepackage[ruled, linesnumbered, vlined]{algorithm2e}

\newcommand{\figref}[1]{Fig.~\ref{#1}}

\newcommand{\equref}[1]{Equ.~(\ref{#1})}

\definecolor{dyellow}{rgb}{0.7, 0.7, 0.0}
\definecolor{ao}{rgb}{0.0, 0.5, 0.0}

\def\rvc{{\mathbf{c}}}

\def\rvx{{\mathbf{x}}}

\def\rvz{{\mathbf{z}}}

\def\rvpi{{\bm{\pi}}}

\def\mA{{\bm{A}}}

\def\mR{{\bm{R}}}
\def\mS{{\bm{S}}}
\def\mT{{\bm{T}}}

\makeatletter
\DeclareRobustCommand\onedot{\futurelet\@let@token\@onedot}
\def\@onedot{\ifx\@let@token.\else.\null\fi\xspace}

\def\eg{\emph{e.g.}~} 
\def\ie{\emph{i.e.}~} 
 
\def\etc{\emph{etc.}~} 
\def\wrt{w.r.t.~} 
\def\etal{\emph{et al.}~}
\makeatother

\newlength\savewidth\newcommand\shline{\noalign{\global\savewidth\arrayrulewidth
		\global\arrayrulewidth 1pt}\hline\noalign{\global\arrayrulewidth\savewidth}}
\newcolumntype{x}[1]{>{\centering\arraybackslash}p{#1pt}}
\newcommand{\tablestyle}[2]{\setlength{\tabcolsep}{#1}\renewcommand{\arraystretch}{#2}\centering\small}

\def\wacvPaperID{xxx} 

\wacvfinalcopy 

\ifwacvfinal
\fi

\ifwacvfinal
\usepackage[breaklinks=true,bookmarks=false]{hyperref}
\else
\usepackage[pagebackref=true,breaklinks=true,colorlinks,bookmarks=false]{hyperref}
\fi

\begin{document}
	
\title{Learning Foreground-Background Segmentation from Improved Layered GANs}

\author{
	Yu~Yang\textsuperscript{1}, ~~%
	Hakan~Bilen\textsuperscript{2}, ~~%
	Qiran~Zou\textsuperscript{1}, ~~%
	Wing~Yin~Cheung\textsuperscript{1}, ~~%
	Xiangyang~Ji\textsuperscript{1}\\
	{\textsuperscript{1} Tsinghua University, BNRist} ~
	{\textsuperscript{2} The University of Edinburgh} \\
	{\tt\small yang-yu16@mails.tsinghua.edu.cn, ~~hbilen@ed.ac.uk}, \\
	{\tt\small \{zouqr19,zhangyx20\}@mails.tsinghua.edu.cn, ~~xyji@tsinghua.edu.cn}
}

\maketitle

\begin{abstract}
Deep learning approaches heavily rely on high-quality human supervision which is nonetheless expensive, time-consuming, and error-prone, especially for image segmentation task. In this paper, we propose a method to automatically synthesize paired photo-realistic images and segmentation masks for the use of training a foreground-background segmentation network. In particular, we learn a generative adversarial network that decomposes an image into foreground and background layers, and avoid trivial decompositions by maximizing mutual information between generated images and latent variables. The improved layered GANs can synthesize higher quality datasets from which segmentation networks of higher performance can be learned. Moreover, the segmentation networks are employed to stabilize the training of layered GANs in return, which are further alternately trained with Layered GANs. Experiments on a variety of single-object datasets show that our method achieves competitive generation quality and segmentation performance compared to related methods.
\end{abstract}

\section{Introduction}

\begin{figure}[t]
	\centering
	\includegraphics[width=\linewidth]{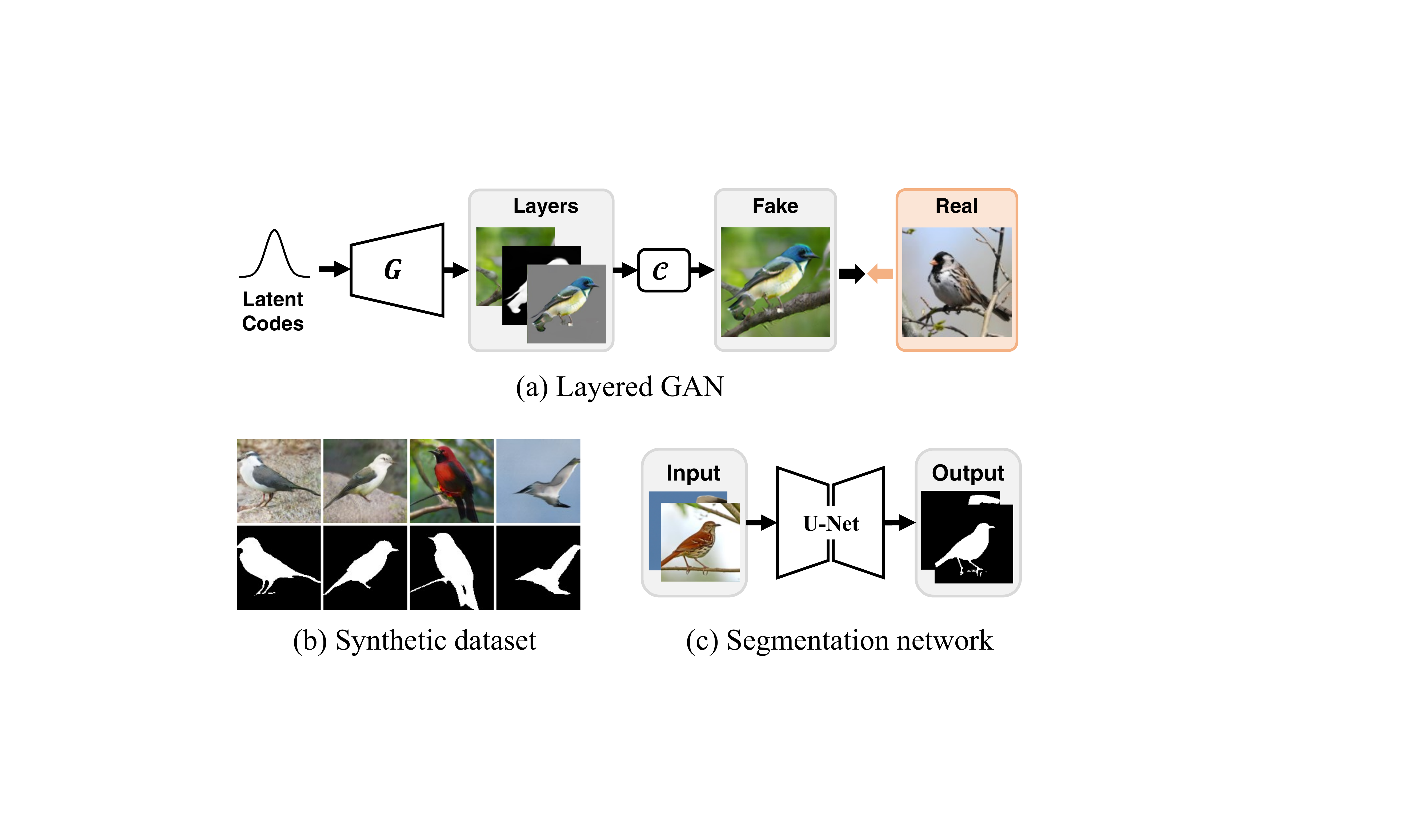}
	\caption{\textbf{Overview}. A layered GAN (a) is trained to synthesize paired image and segmentation masks (b) for training a segmentation network, \eg U-Net (c). $\mathcal{C}$ denotes the composition module.}
	\label{fig:teaser}
\end{figure}

Deep learning approaches continue to dramatically push the state-of-the-art in most supervised computer vision tasks, but they demand tremendous number of human annotations which are expensive, time-consuming, and error-prone, especially for image segmentation task.
Hence there is a growing interest in reducing the reliance on human supervision.
Towards this target, we investigate learning to synthesize paired images and segmentation masks without segmentation annotation for the use of training segmentation network (\figref{fig:teaser}).

\begin{figure}[t]
	\centering
	\includegraphics[width=\linewidth]{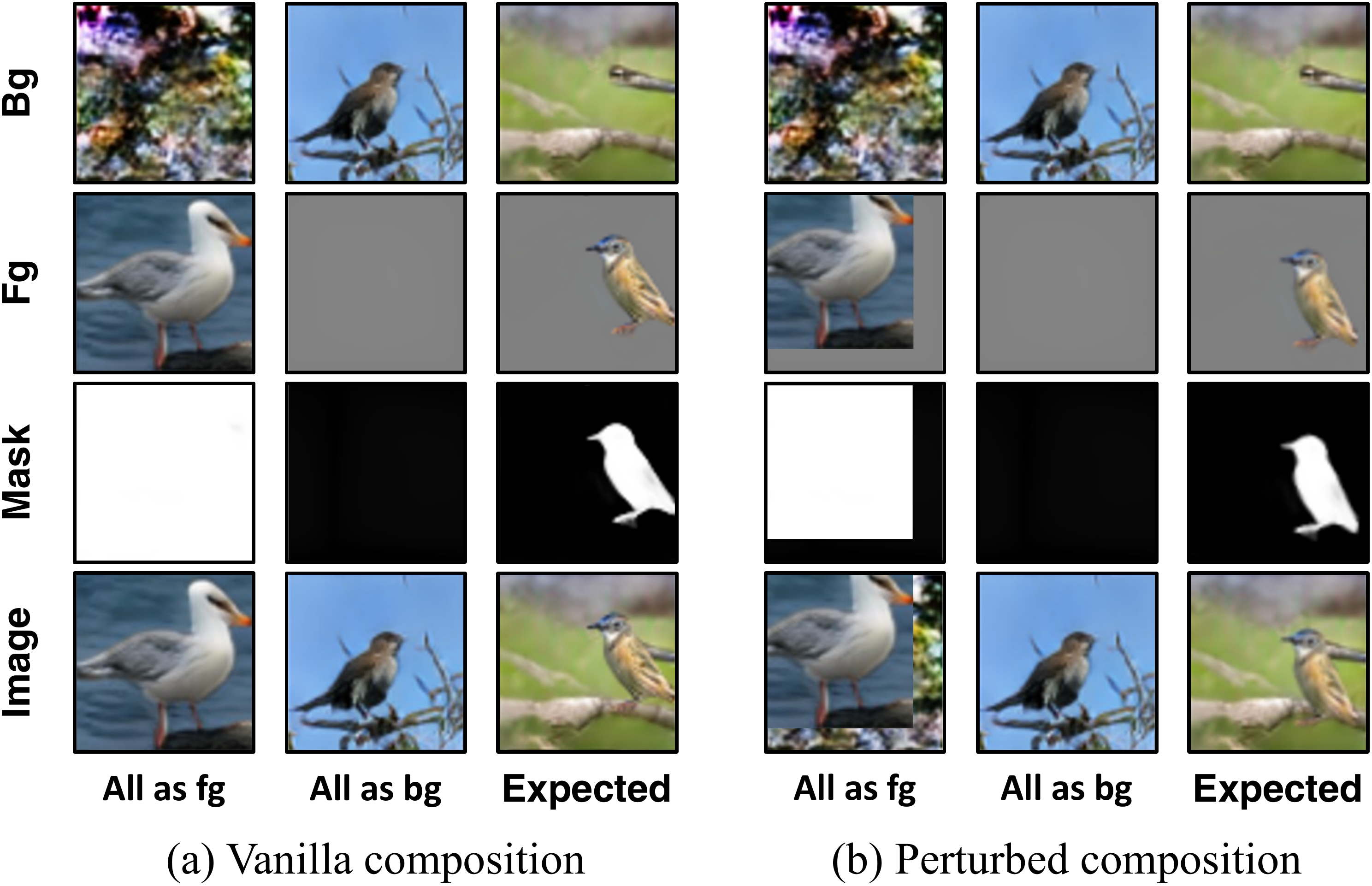}
	\caption{\textbf{Illustrations} of the learned composition and the effect of perturbation. Vanilla composition permits ``all as foreground'' ($\rvpi=\bm{1}$), ``all as background'' ($\rvpi=\bm{0}$), and expected composition. Perturbed composition penalizes ``all as foreground'' by causing unnatural border, maintains the fidelity for expected composition, but is unable to counter ``all as background''.}
	\label{fig:trivial}
\end{figure}

In the recent years, generative adversarial networks (GANs)~\cite{goodfellow2014generative} repeatedly show its ability to not only synthesize highly photorealistic images~\cite{karras2019style, karras2020analyzing, brock2018large} but also disentangle attributes~\cite{karras2019style, karras2020analyzing} and geometric factors~\cite{zhang2020image}.
To make GANs factorize image layers (\ie foreground and background), a body of works~\cite{yang2017lr, singh2019finegan, benny2020onegan, chen2019unsupervised, bielski2019emergence}, dubbed \emph{layered} GANs in this paper, employ a compositional generation process which first generates background and foreground objects and then composes them to obtain the synthetic image.
This compositional generation process, however, does not guarantee an expected factorization and is prone to trivial compositions such as ``all as foreground'' and ``all as background'' (\figref{fig:trivial}(a) and \figref{fig:ablation_mi}) without any supervision or any regularization.

PerturbGAN~\cite{bielski2019emergence} addresses this issue by perturbing the foreground position during composition, as such perturbation leads to inferior generated images for all-as-foreground decomposition and maintains decent generated images for expected decomposition (\figref{fig:trivial}(b)). 
Despite so, perturbation is unable to solve the all-as-background issue, or empty foreground issue, where it can not affect the fidelity of generated image.
To alleviate this problem, PerturbGAN~\cite{bielski2019emergence} penalizes the cases where the number of foreground pixels are smaller than a threshold. 
However, this method is not always effective and the additionally introduced hyperparameter  is sensitive to object scale, object category and dataset.

We propose a GAN that synthesizes background and foreground from a pair of latent codes.
In more details, one of the latent codes, dubbed public code, is shared across background and foreground generation, while the other, dubbed private code, is only reserved for foreground generation.
In contrast to PerturbGAN~\cite{bielski2019emergence}, we not only employ perturbation and require our generators to generate photorealistic images but also diversify the generated foreground by maximizing the mutual information between the composed image and the private code as well as the mutual information between the foreground mask and the private code.
To this end, the all-as-background issue can be effectively mitigated and the layered GANs decompose foreground and background in a superior way.

Our method is reminiscent of a rich body of work~\cite{eslami2016, burgess2019monet, greff2019multi, engelcke2019genesis, lin2020space, engelcke2021genesis, anciukevicius2020object} on object-centric scene generation, which is mostly based on variational autoencoders (VAEs).
These models are able to decompose and generate multi-instance complex scenes.
Nonetheless, these methods are mostly demonstrated on synthetic scene but seldom applied to real-world images \cite{engelcke2021genesis}.
In contrast, our method has an advantage in dealing with realistic images but lack the ability to model multi-instance generation.

Our method presents a solution to unsupervised foreground-background segmentation (\figref{fig:teaser}) where layered GANs are trained to generated synthetic training data from which segmentation networks are trained. 
In return, the segmentation networks are used to further regularize the generation of layered GANs.
These two steps are alternated which finally leads to a high-performance segmentation network.
Our method learns segmentation with the analysis-by-synthesis principle, which is different from earlier work~\cite{rother2006cosegmentation, hochbaum2009efficient, joulin2010discriminative, alexe2010classcut, chai2011bicos} that learns to discriminate foreground and background pixels using shallow classifiers on hand-crafted features and the recent trend to simultaneously learn per-pixel representations and clustering \cite{ji2019invariant, ouali2020autoregressive, kim2020unsupervised} with deep neural networks. 

We evaluate our method on a variety of single-object datasets including Caltech-UCSD Birds 200-2011 (CUB), Stanford Cars, Stanford Dogs, and Amazon Picking Challenge (APC), where it outperforms the baseline methods and achieves competitive performance compared to related methods.

Our contributions are summarized as follows:

1) We propose a mutual information based learning objective and show that it achieves significantly better performance in terms of disentangling foreground and background.

2) We show that the synthesized image and segmentation mask pairs can be successfully used to train segmentation networks which achieve competitive performance compared to related methods.


\section{Related Work}\label{sec:related_works}
\paragraph{Layered GANs}
Generative adversarial networks (GANs)~\cite{goodfellow2014generative} have been proposed and improved towards highly photorealistic image synthesis~\cite{karras2019style, brock2018large, karras2020analyzing} and explored for disentangling factors of variation towards controllable generation~\cite{goetschalckx2019ganalyze, shen2020interpreting, karras2019style, plumerault2019controlling, jahanian2019steerability, voynov2020unsupervised, spingarn2020gan, harkonen2020ganspace}, efficient representation learning~\cite{donahue2019large}, and saving of human annotations~\cite{zhang2021datasetgan, zhang2020image}.
To gain controllable generation at object level, layered GANs~\cite{yang2017lr,singh2019finegan, bielski2019emergence, benny2020onegan} leverage the compositional nature of images and synthesize images by composing image layers, \ie background and foreground objects.
This compositional generation process is further developed to support 3D-aware scene generation~\cite{liao2020towards, BlockGAN2020, niemeyer2021giraffe, niemeyer2021campari} and video generation~\cite{relate2020}.
Unlike most work that is interested in generation quality, our work focus on improving and employing 2D layered GANs towards learning segmentation with minimal human supervision.

In layered GANs, it is observed that compositional generation process is prone to degeneration to single layer generation \cite{yang2017lr, chen2019unsupervised, bielski2019emergence}.  This issue is alleviated by composing layers at feature level~\cite{relate2020, niemeyer2021giraffe}, using weak supervision (\eg bounding box annotation or background images)~\cite{liao2020towards, singh2019finegan, benny2020onegan}, or specific regularization (\eg perturbation strategy as in PerturbGAN~\cite{bielski2019emergence}).
We propose a method based on mutual information maximization to mitigate this issue which can be successfully employed in layered GANs under the challenging unsupervised learning setting.

\paragraph{Object-Centric Scene Generation}
A rich body of work~\cite{eslami2016, burgess2019monet, greff2019multi, engelcke2019genesis, lin2020space, engelcke2021genesis, anciukevicius2020object}, grounded on variational autoencoders (VAEs)~\cite{kingma2014auto}, simultaneously learn scene decomposition, object-centric representations, inter-object relationships, and multi-instance scene generation.
Although these works also employ a compositional generation process and interested in the automatic segmentation, they are mostly demonstrated on synthetic scenes but rarely evaluated on real-world datasets. 
In contrast, our work focuses on compositional generation and segmentation of real-world datasets. 

\paragraph{Unsupervised Segmentation} 
Earlier works~\cite{rother2006cosegmentation, hochbaum2009efficient, joulin2010discriminative, alexe2010classcut, chai2011bicos} cast unsupervised object segmentation as co-segmentation and utilise handcrafted features and shallow classifiers. 
With the advent of deep learning, per-pixel representation and clustering are simultaneously learned with deep neural networks \cite{ji2019invariant, ouali2020autoregressive, kim2020unsupervised} to achieve unsupervised segmentation. 
Apart from these works, other works learn segmentation from generative models.
In particular, segmentation hint is extracted via cut and paste~\cite{remez2018learning, arandjelovic2019object, bielski2019emergence, abdal2021labels4free}, manipulating generation process~\cite{voynov2020big, melas2021finding}, erasing and redrawing~\cite{chen2019unsupervised}, or maximizing inpainting error~\cite{savarese2020information}. 
Our methods are based on generative models and learn segmentation from compositional generation process.


\section{Methods}\label{sec:methods}

\subsection{Revisiting Layered GANs}
Let $G:\mathcal{Z}\to\mathcal{X}$ denote a generative model that maps a latent variable $\rvz\in\mathcal{Z}$ to an image $\rvx\in\mathcal{X}$.
In GANs, there is also a discriminator $D:\mathcal{X}\to\mathbb{R}$ tasked to classify real and fake images. 
$G$ and $D$ are trained by playing the following adversarial game
\begin{equation}
	\min_{G} \max_{D}~~\mathbb{E}_{\rvx\sim p_{\text{data}}(\rvx)} f(D(\rvx)) 
	+ \mathbb{E}_{\rvz} f(-D(G(\rvz))),
\end{equation}
where $f:\mathbb{R}\to\mathbb{R}$ is a concave function which differs for different kinds of GANs. In the end of adversarial learning, the images generated from $G$ are high-fidelity and is hardly distinguished by $D$.

To explicitly disentangle foreground objects from background, layered GANs decompose $G$ into multiple generators responsible for generating layers which are further composed to synthesize full images. Formally, regarding two-layer generation, $G$ is decomposed into background generator $G_b$ and foreground generator $G_f$ as $G = (G_b, G_f)$.
$G_b$ generates background image $\rvx_b$ and $G_f$ generates paired foreground image $\rvx_f$ and mask $\rvpi$. 
The synthetic image $\rvx$ is completed by blending $\rvx_f$ and $\rvx_b$ with $\rvpi$ as alpha map,
\begin{equation}
	\label{eq:superimpose}
	\rvx = \mathcal{C}(\rvx_b, \rvx_f, \rvpi) = (1 - \rvpi) \odot \rvx_b + \rvpi \odot \rvx_f,
\end{equation}
where $\mathcal{C}$ denote the composition step.
However, this naive layered GAN is prone to trivial decomposition such as ``all as foreground'' and ``all as background'' (\figref{fig:trivial} and \figref{fig:ablation_mi}), which requires further regularization as follows. 

\paragraph{Layer perturbation} 
Bielski \& Favaro \cite{bielski2019emergence} propose to perturb the position of foreground object in composition step during training.
Formally, this perturbed composition can be formulated as 
\begin{equation}\label{eq:perturbation}
	\mathcal{C}(\rvx_b, \rvx_f, \rvpi; \mathcal{T}) = 
	(\bm{1} - \mathcal{T} (\rvpi)) \odot \rvx_b +
	\mathcal{T}(\rvpi) \odot \mathcal{T} (\rvx_f),
\end{equation}
where $\mathcal{T}$ denotes a perturbation operator which
PerturbGAN \cite{bielski2019emergence} instantiates as pixel-unit translation. In this paper, we employ restricted affine transformation. 
This method can effectively alleviate the all-as-foreground issue, because artefacts like sharp and unnatural border (\figref{fig:trivial}(b)) emerges in the images under this circumstance and such inferior generation is penalized by discriminator.
It should be noted that this perturbed composition does not hurt the fidelity of composed images in expected decomposition as long as the perturbation is imposed within a small range, because the relative position of foreground with respect to background can be varied up to a certain degree for most images.
However, the perturbed composition is insufficient to address the all-as-background issue, which brings to our mutual information maximization as in the next section.

\subsection{Improving Layered GANs with Mutual Information Maximization}\label{sec:mutual_information}

\begin{figure*}[t]
	\includegraphics[width=\linewidth]{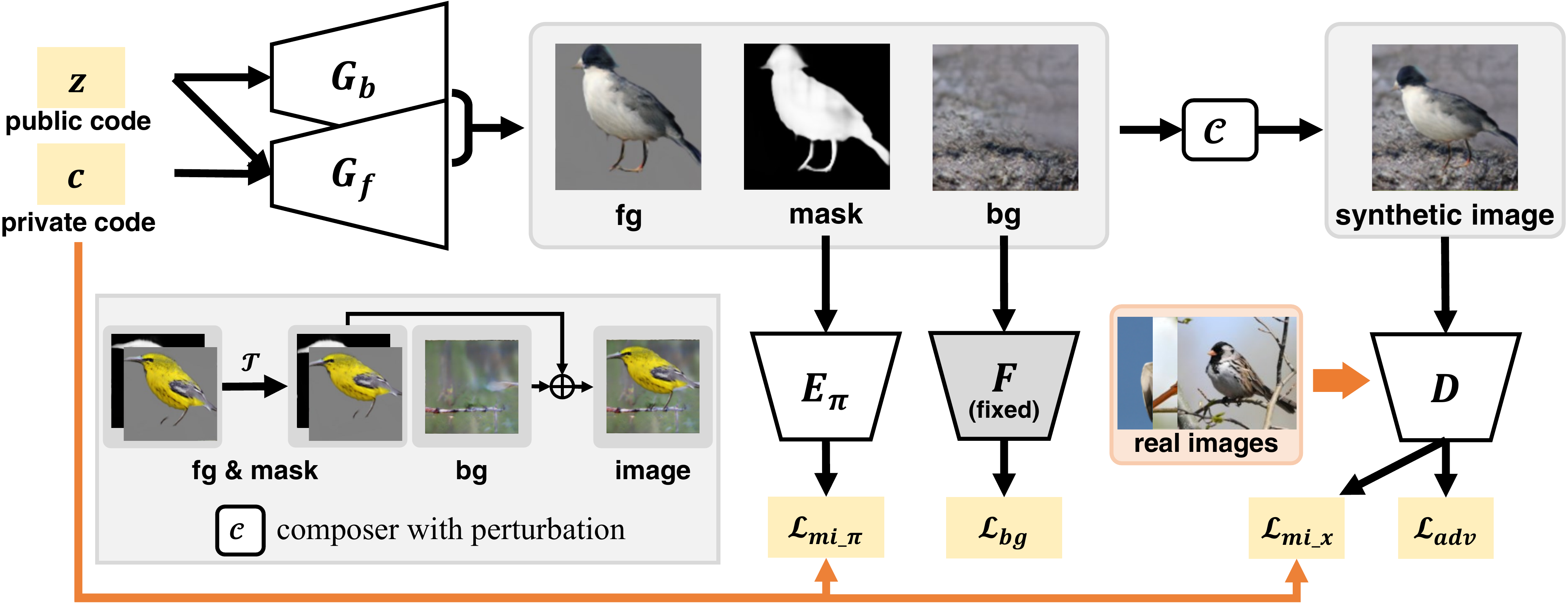}
	\caption{\textbf{Framework}. Our generative model comprises background generator $G_b$, foreground generator $G_f$, and composer with perturbation $\mathcal{C}$. The public code $\rvz$ is shared across foreground and background generation, while the private code $\rvc$ is only used for foreground generation.
		Our generative model is mainly trained via adversarial learning against a discriminator $D$, denoted as $\mathcal{L}_\text{adv}$ and regularized by mutual information loss, denoted as $\mathcal{L}_{\text{mi}\_\pi}$ and $\mathcal{L}_{\text{mi}\_x}$, whose estimation is assisted by $E_\pi$ and $E_x$. $E_x$, which is omitted in the figure, shares backbone parameters with $D$ and bifurcates at the top layers. In the alternate training, background generation is regularized by segmentation network $F$, denoted as $\mathcal{L}_\text{bg}$.}
	\label{fig:framework}
\end{figure*}

\figref{fig:framework} presents the framework of our model. 
Generally, we improve layered GANs with in terms of disentangling foreground object and background from a special design of generative model and mutual information maximization, which is described as follows.

Our generative model takes as input two kinds of latent code, denoted as $\rvz$ and $\rvc$, respectively. 
In particular, $\rvz$ is shared across $G_b$ and $G_f$, whereas $\rvc$ is private to $G_f$. We therefore refer to $\rvz$ and $\rvc$ as public code and private code, respectively (\figref{fig:framework}). 
Based on this design, we introduce an additional learning objective aiming at maximizing mutual information between the composed image $\rvx$ and the private code $\rvc$ and mutual information between the generated mask $\rvpi$ and the private code $\rvc$, 
\begin{equation}\label{eq:mutual_information}
	\max I(\rvx, \rvc) + I(\rvpi, \rvc).
\end{equation}
This learning objective can penalize $G_f$ from always generating zero mask, which is shown as following.

For the first term in \equref{eq:mutual_information}, it is the composed image $\rvx$ that is measured the mutual information with $\rvc$. If $G_f$ produces all-zero mask, \ie $\rvpi=\bm{0}$, the composed image would only contain background, \ie $\rvx=\rvx_b$, and $I(\rvx, \rvc)= I(\rvx_b, \rvc) = 0$ since $\rvc$ is private to foreground generation process.
Therefore, non-zero mask is essential for maximizing $I(\rvx, \rvc)$. For the second term, the mutual information can be rewritten as difference between two entropies, $I(\rvpi, \rvc) = H(\rvpi) - H(\rvpi|\rvc)$. Note that our generative model is deterministic, meaning there is no stochastic in $\rvpi$ given $\rvc$, \ie $H(\rvpi|\rvc)=0$. Hence, maximizing $I(\rvpi, \rvc)$ is equivalent to maximizing $H(\rvpi)$. If $G_f$ constantly generates all-zero masks, $H(\rvpi)$, which depicts the diversity of masks, would be very low. Therefore, maximizing $I(\rvpi, \rvc)$ also advocates non-zero masks.

In practice, computing the mutual information, $I(\rvx, \rvc)$ and $I(\rvpi, \rvc)$, is intractable. Following InfoGAN \cite{chen2016infogan}, we instead maximize the variational lower bound of the mutual information, which is realized as following. First, neural networks $E_\rvpi$ and $E_\rvx$ are introduced to approximate the posterior distribution $p(\rvc|\rvpi)$ and $p(\rvc|\rvx)$, respectively. Second, $G$ and $E_\rvpi$ and $E_\rvx$ are learned by minimizing the following loss
\begin{equation}\label{eq:info_loss}
	\mathcal{L}_{mi} = \mathbb{E}_{\rvz, \rvc} \log E_\rvx(\rvc|\rvx) + \log E_\rvpi(\rvc|\rvpi)
\end{equation}
where $E_\rvx(\rvc|\rvx)$ and $E_\rvpi(\rvc|\rvpi)$ denote the likelihood of a sample $\rvc$ under the approximate posterior distribution of $p(\rvc|\rvpi)$ and $p(\rvc|\rvx)$, respectively.
At the implementation level, $E_\rvx$ shares backbone network parameters with discriminator $D$ and bifurcates at the last two layers, as commonly done in~\cite{chen2016infogan, lin2020infogan, singh2019finegan, benny2020onegan}. More implementation details can be found in the appendices. 

In addition to penalizing trivial masks, introducing private code and maximizing $I(\rvx, \rvc)$ is also favorable for \emph{foreground-oriented clustering}. In particular, $\rvc$ can be designed as discrete variable and its prior distribution can be a uniform categorical distribution. In this way, $\rvc$ is interpreted as clustering category and $E_\rvx$ is regarded as a clustering network trained from synthetic data. 
Note that the unique design of our generative model, where $\rvc$ is only used for foreground generation, makes the clustering based only on the variety of foreground objects and not interfered by backgrounds. 
This mechanism conforms to the logics of category annotation for major datasets, where it is foreground objects that are categorized rather than background. 
Experimental results in Section~\ref{sec:fine_grained_results} (Table~\ref{tab:quantitative_fine_grain}) justifies the merit of our foreground-oriented clustering.

\subsection{Alternate Training of Layered GAN and Segmentation Network}

\paragraph{Training a layered GAN}
The improved Layered GAN is optimized by updating generator and discriminator alternately,
\begin{equation}\label{eq:lgan}
	\begin{aligned}
		& \min_{G} \mathcal{L}_\text{adv}^{(G)} 
		+ \gamma_\text{mi} \mathcal{L}_\text{mi}^{(G)}
		+ \gamma_\text{b} \mathcal{L}_\text{b}^{(G)}, ~~~~\\
		& \min_{D, E_\rvx, E_\rvpi} \mathcal{L}_\text{adv}^{(D)} + 
		\gamma_\text{mi} \mathcal{L}_\text{mi}^{(E_\rvx, E_\rvpi)},
	\end{aligned}
\end{equation}
where $\mathcal{L}_\text{mi}$ denotes the mutual information loss as defined in \equref{eq:info_loss}, $\mathcal{L}_\text{adv}$ denotes the adversarial loss for which hinge loss is employed, \ie 
$\mathcal{L}_\text{adv}^{(G)} =   
- \mathbb{E}_{\rvz, \rvc} D(G(\rvz, \rvc))$,
$\mathcal{L}_\text{adv}^{(D)} = 
\mathbb{E}_{\rvx\sim p_{\text{data}}(\rvx)} \max(0, 1 - D(\rvx)) + \mathbb{E}_{\rvz} \max(0, 1 + D(G(\rvz, \rvc))),$
$\mathcal{L}_b^{(G)}$ denotes the binarization loss 
$\mathcal{L}_b^{(G)} = \mathbb{E}_{\rvz, \rvc} \min(\rvpi, \bm{1} - \rvpi)$
to encourage the binarization of masks as in~\cite{bielski2019emergence}, and $\gamma_\text{mi}$, and $\gamma_b$ denote the corresponding loss weights.

\paragraph{Training a Segmentation Network}\label{sec:unsupseg}
Improved layered GAN can synthesize paired images and segmentation masks by sampling latent variables from prior distribution and forwarding them to the generator.  
This procedure is repeated for $N$ times to generate a synthetic dataset $\mathcal{D}_{syn}=\{(x_i, \pi_i)\}_{i=1}^N$ which is further used for training a foreground-background segmentation network  $F:\mathcal{X}\rightarrow[0,1]^{H\times W}$ as follows,
\begin{equation}\label{eq:train_seg}
	\min_{F} \mathbb{E}_{(x, \pi)\sim\mathcal{D}_{syn}}
	[(1 - \pi) \log(1 - F(x)) + \pi \log(F(x))].
\end{equation}
Since the layered GAN approximates the real data distribution and produces reasonable masks, the optimized $F$ could fairly generalize to real images.

\begin{algorithm}[t]\label{alg:alternate}
	\caption{Alternate training}
	\For {$k = 1, \cdots, K$} {
		Set $\hat{\gamma}_\text{bg} = 0$ if $k=1$ else $\hat{\gamma}_\text{bg}=\gamma_\text{bg}$
		
		\For {$n = 1, \cdots, N_{gan}$} {
			Update $G_b$ and $G_f$ with loss 
			$\mathcal{L}^{(G)} = \mathcal{L}_\text{adv}^{(G)} 
			+ \gamma_\text{mi} \mathcal{L}_\text{mi}^{(G)}
			+ \gamma_\text{b} \mathcal{L}_\text{b}^{(G)}
			+ \hat{\gamma}_\text{bg} \mathcal{L}_\text{bg}^{(G)}$
			
			Update $D$, $E_x$ and $E_\pi$ with loss
			$\mathcal{L}^{(D)} = \mathcal{L}_\text{adv}^{(D)} + 
			\gamma_\text{mi} \mathcal{L}_\text{mi}^{(E_\rvx, E_\rvpi)}$
		}
		
		\For {$n = 1, \cdots, N_{seg}$} {
			Generate a minibatch of synthetic dataset $\mathcal{B}$.
			
			Update $F$ on $\mathcal{B}$ with loss 
			$\mathcal{L}^{(F)} = \mathbb{E}_{(x, \pi)\sim\mathcal{B}}
			[(1 - \pi) \log(1 - F(x)) + \pi \log(F(x))]$  
		}
		
	}
\end{algorithm}

\paragraph{Alternate Training} 
As presented in Algorithm~\ref{alg:alternate}, an alternate training schedule is employed to exploit the synergy between layered GAN and segmentation segmentation network. In particular, the training is conducted for $K$ rounds. In the initial round, a layered GAN is trained without the help of segmentation network. The initially learned layered GAN is sufficient to generate synthetic dataset for training a segmentation network.
In the rest rounds, the segmentation network is used to regularize the background generation with an additional loss term
\begin{equation}\label{eq:loss_bg}
	\mathcal{L}_\text{bg}^{(G)} = \mathbb{E}_{\rvz, \rvc} \log(1 - F(\rvx_b)),
\end{equation}
where the parameters of segmentation network are fixed. 
This loss encourages the generated background pixels, $\rvx_b$, to be classified as background and helps to stabilize the learned disentanglement of foreground and background.
The training of layered GAN is then followed by another time of updating segmentation network. 
In this way, layered GAN and segmentation network are alternately trained.
More training details are available in the appendices.


\section{Experiments}\label{sec:experiments}
\subsection{Settings}
\paragraph{Datasets} 
Our method is evaluated on a variety of datasets including \textit{Caltech-UCSD Birds 200-2011} (CUB)~\cite{WahCUB_200_2011}, \textit{Stanford Cars}~\cite{krause20133d}, \textit{Stanford Dogs}~\cite{khosla2011novel}, and \textit{Amazon Picking Challenge} (APC)~\cite{zeng2016multi}. CUB, Stanford Cars and Stanford Dogs are fine-grained category datasets for bird, car, and dog, respectively. Most images in these three datasets only contain single object.
APC is a image dataset collected from robot context which contains images of single object either on a shelf or in a tray. Unlike the above four datasets, objects in APC belong to a variety of categories such as ball, scissors, book, \etc
Ground truth segmentation masks of CUB and APC are available from their public release, whereas pre-trained Mask R-CNN are employed to approximate the ground truth segmentation of Stanford Cars, and Stanford Dogs, as in previous work~\cite{singh2019finegan, benny2020onegan}.
More details about the datasets are available in the appendices.

\paragraph{Evaluation metrics} 
Our methods are quantitatively evaluated with respect to generation quality and segmentation performance.
To measure the quality of generation, Fréchet Inception Distance (FID)~\cite{heusel2017gans} between $10$k synthetic images and all the images are computed.
The segmentation performance is evaluated with per-pixel accuracy (ACC), foreground intersection over union (IoU) between predicted masks and ground truth ones, and mean IoU (mIoU) across foreground and background.
Moreover, adjusted random index (ARI)~\cite{greff2019multi} and mean segmentation covering (MSC)~\cite{engelcke2019genesis} are computed to compare the segmentation performance level of our method to that of VAE-based methods.
In addition to above metrics, normalized mutual information (NMI) is computed to measure the category clustering performance of our method which is compared to other layered GANs.

\subsection{Ablation Studies}
We conduct multiple ablation studies to investigate (i) the effect of mutual information maximization, (ii) the influence of different choices for private code $\rvc$ distribution, and (iii) the effect of alternate training by controlled experiments which are mainly conducted in CUB dataset at $64\times64$ resolution.

\paragraph{The effect of mutual information maximization}
\begin{figure*}[t]\centering
	\includegraphics[width=0.95\linewidth]{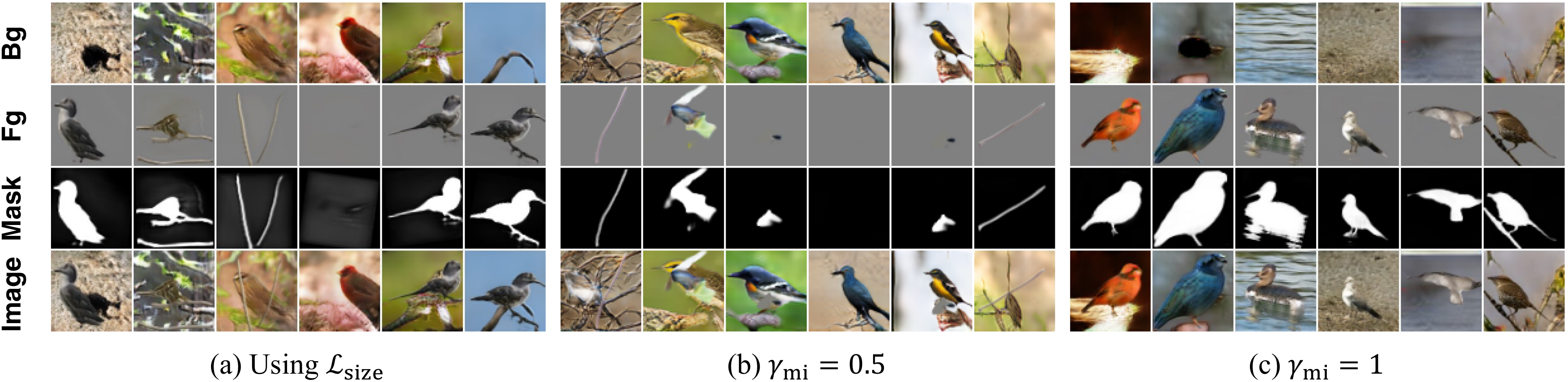}
	\caption{\textbf{Ablation study on mutual information maximization} on CUB. $\gamma_\text{mi}$ denotes the loss weight of mutual information maximization. $\mathcal{L}_\text{size}$ denotes that mutual information maximization is replaced with with mask size loss.}
	\label{fig:ablation_mi}
\end{figure*}

\begin{table}[t]\centering
	\tablestyle{7pt}{1.25}\begin{tabular}{c|c|ccccc}
		\shline
		& \multirow{2}{*}{$\mathcal{L}_\text{size}$} & \multicolumn{5}{c}{$\gamma_\text{mi}$} \\
		& 	& 0.1 & 0.5 & 1.0 & 5.0 & 10.0 \\
		\hline	
		FID~$\downarrow$ 	& 19.4 & 19.1 & 17.6 & \textbf{11.6} & 15.5 & 17.8 \\
		bg-FID~$\uparrow$ 	& 20.9 & 18.6 & 17.4 & 99.5 & 107.1 & \textbf{120.2} \\
		IoU~$\uparrow$ 		& 6.3 & 0.	& 0.1	& \textbf{68.6} & 66.5 & 66.3 \\
		\shline
	\end{tabular}
	\caption{\textbf{Ablation study with respect to mutual information maximization} on CUB. $\gamma_\text{mi}$ denotes the loss weight of mutual information maximization. $\mathcal{L}_\text{size}$ denotes the mask size loss.}
	\label{tab:ablation_mi}
\end{table}

Table~\ref{tab:ablation_mi} and \figref{fig:ablation_mi} presents the results of ablation study with respect to mutual information maximization, where $\gamma_{mi}$ is varied or mask size loss $\mathcal{L}_\text{size}$ is used to replace mutual information maximization. In addition to FID and IoU, we also compute bg-FID which is the FID between synthetic background image and realistic images. To some extent, a bg-FID that is close to FID can manifest that ``all as background'' occurs. Therefore, we would expect there is sufficient gap between bg-FID and FID. 
It can be seen from \figref{fig:ablation_mi} that ``all as background'' still occurs when using mask size loss. This is also evidenced by the poor segmentation performance and the negligible gap between FID and bg-FID (Table~\ref{tab:ablation_mi}). 
In contrast, the utilization of mutual information loss with appropriate weight (\eg $\gamma=1, 5, 10$) can successfully suppress the occurrence of ``all as background''.
It is noteworthy that the benefits from mutual information loss saturates and the FID score, which reflects the fidelity and diversity of synthetic images, drops as $\gamma_\text{mi}$ increases beyond 1.0.
We attribute this to adversarial learning being overwhelmed by mutual information maximization. To trade off between the quality of  synthetic fidelity and suppression of degeneration, a proper $\gamma_\text{mi}$ needs to be chosen, but the secure range of $\gamma_\text{mi}$ is quite wide as the performances of $\gamma_\text{mi}$ from 1 to 10 are quite close. Additional results on other datasets are available in the supplementary material.

\begin{table}[t]\centering
	\tablestyle{5pt}{1.25}\begin{tabular}{c|cccc|cccc}
		\shline
		& \multicolumn{4}{c|}{Normal} & \multicolumn{4}{c}{Categorical}\\
		$d_c$ 	& 4 & 6 & 8 & 10 & 50 & 100 & 200 & 400 \\
		$H(\rvc)$ & 5.6 & 8.5  & 11.3 & 14.1 & 3.9 & 4.6 & 5.2 & 5.9 \\
		\hline	
		FID 	& 12.3 & 12.1 & 11.8 & 11.1 & 11.3 & 11.6 & 11.6 & 11.9 \\
		IoU 	& 72.3 & 72.5 & 66.7 & 65.4 & 71.5 & 69.1 & 68.6 & 68.6 \\
		\shline
	\end{tabular}
	\caption{\textbf{Comparison of different prior distribution for $\rvc$} in CUB. $d_c$ denotes the dimension of $\rvc$ for normal distribution and the number of categories for categorical distribution. $H(\rvc)$ denotes the entropy of $\rvc$ drawn from the prior distribution.}
	\label{tab:ablation_c_distribution}
\end{table}

\paragraph{Prior distribution of $\rvc$}
We investigate the influence of different choices for prior distribution of $\rvc$. In particular, normal distribution and categorical distribution are compared and the dimension (or number of categories) of $\rvc$ is varied. Results in Table~\ref{tab:ablation_c_distribution} show that our method achieves similar performance with different settings, suggesting that the proposed mutual information maximization is robust to different choices for prior distribution of $\rvc$. Although normal distribution leads to marginally higher performance than categorical distribution, the categorical distribution advocates the foreground-oriented clustering as discussed in~\ref{sec:mutual_information} and makes our method comparable to other layered GANs~\cite{singh2019finegan, benny2020onegan}.

\begin{table}[t]\centering
	\tablestyle{5pt}{1.25}\begin{tabular}{l|cc|cc|cc}
		\shline
		 & \multicolumn{2}{c|}{CUB} & \multicolumn{2}{c|}{Stanford Dogs} & \multicolumn{2}{c}{Stanford Cars} \\
		& FID~$\downarrow$ & IoU~$\uparrow$ & FID~$\downarrow$ & IoU~$\uparrow$ & FID~$\downarrow$ & IoU~$\uparrow$ \\
		\hline
		w/o A.T. & \textbf{11.6} & 68.6 & 67.5 & 53.2 & 13.0 & 58.3 \\
		w/ A.T. & 12.9  & \textbf{70.1} & \textbf{59.3} & \textbf{61.3} & \textbf{12.8} & \textbf{62.0} \\	
		\shline
	\end{tabular}
	\caption{\textbf{Ablation study with respect to alternate training} on CUB, Stanford Dogs, and Stanford Cars. A.T. denotes alternate training.}
	\label{tab:ablation_alternate}
\end{table}

\paragraph{The effect of alternate training}
We compare the performance of our methods with and without the alternate training strategy on CUB, Stanford Dogs, and Stanford Cars at $64\times64$ resolution. Table~\ref{tab:ablation_alternate} shows that the alternate training helps to improve the segmentation performance without compromising the generation quality.

\subsection{Results in Fine-Grained Category Datasets}\label{sec:fine_grained_results}
\begin{figure*}[t]\centering
	\includegraphics[width=\linewidth]{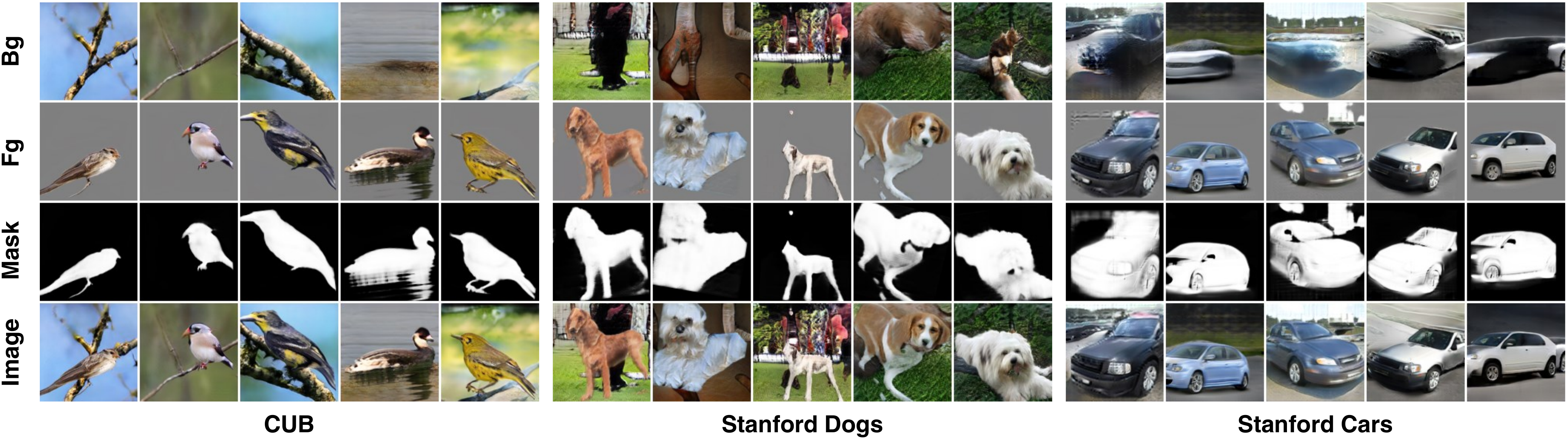}
	\caption{\textbf{Qualitative generation results} in fine-grained category datasets: CUB, Stanford Dogs, Stanford Cars. From top to bottom, the background (Bg), foreground multiplied with mask (Fg), foreground mask (Mask), and composed image (Image) are presented.}
	\label{fig:qualitative_gen}
\end{figure*}

\begin{table*}[t]\centering
	\tablestyle{9.5pt}{1.25}\begin{tabular}{lc|ccc|ccc|ccc}
		\shline
		\multirow{2}{*}{Methods} & \multirow{2}{*}{Sup.} & \multicolumn{3}{c|}{CUB} & \multicolumn{3}{c|}{Stanford Dogs} & \multicolumn{3}{c}{Stanford Cars} \\
		& & FID~$\downarrow$ & IoU~$\uparrow$ & NMI~$\uparrow$ 
		& FID~$\downarrow$ & IoU~$\uparrow$ & NMI~$\uparrow$ 
		& FID~$\downarrow$ & IoU~$\uparrow$ & NMI~$\uparrow$ \\
		\hline
		FineGAN \cite{singh2019finegan} & Weak & 23.0 & 44.5 & .349 & 54.9 & 48.7 & .194 & 24.8 & 53.2 & .233 \\
		OneGAN \cite{benny2020onegan}	& Weak & 20.5 & 55.5 & .391 & \textbf{48.7} & \textbf{69.7} & .211 & 24.2 & \textbf{71.2} & .272 \\
		\hline
		Ours	& Unsup. & \textbf{12.9} & \textbf{69.7} & \textbf{.585} & 59.3 & 61.3 & \textbf{.257}  & \textbf{19.0} & 63.3 & \textbf{.301} \\	
		\shline
	\end{tabular}
	\caption{\textbf{Comparison to other layered GANs} in fine-grained category datasets: CUB, Stanford Dogs, and Stanford Cars. Generation quality, segmentation performance, and clustering performance are evaluated with FID, IoU, and NMI, respectively. Note that FineGAN and OneGAN is trained under weakly supervised setting, while our model is trained under unsupervised setting.}
	\label{tab:quantitative_fine_grain}
\end{table*}

Results of our method in three fine-grained category datasets, CUB, Stanford dogs, and Stanford Cars, are presented and compared to that of other layered GANs, FineGAN~\cite{singh2019finegan} and OneGAN~\cite{benny2020onegan}.
All of the experiments are conducted at $128\times128$ resolution.
It should be noted that FineGAN and OneGAN requires all of or a portion of training images annotated with bounding boxes, which should be regarded as weak supervision. 
On the contrary, our models only learn from images, which is should be thought of as unsupervised learning.
Table~\ref{tab:quantitative_fine_grain} compares our method to FineGAN and OneGAN in the aspect of generation quality, segmentation performance, and clustering performance with FID, IoU, and NMI, respectively, and \figref{fig:qualitative_gen} presents the qualitative generation results of our method. 

Despite in the more challenging unsupervised learning setting, our method is still able to learn disentangled foreground and background and synthesize quality paired images and segmentation masks (\figref{fig:qualitative_gen}).
Our method even achieves higher generation quality than FineGAN and OneGAN in CUB and Stanford Cars, with 7.6 and 5.2 higher FID, respectively.
The segmentation performance of our method is still competitive to that of FineGAN and OneGAN. It is notable that our method even outperforms OneGAN in CUB with large margin (14.2 IoU) and achieves close performance to OneGAN on Stanford Dogs and Stanford Cars.
We also take the trained $E_x$ as a clustering network and evaluate its category clustering performance (Table~\ref{tab:quantitative_fine_grain}). It shows that our method constantly obtains higher NMI score than FineGAN and OneGAN. This result justifies the merit of our foreground-oriented clustering as the ground truth category label is annotated based on foreground rather than background as discussed in Section~\ref{sec:mutual_information}.

\subsection{Object Segmentation Results}
\begin{figure*}[t]
	\centering
	\includegraphics[width=\linewidth]{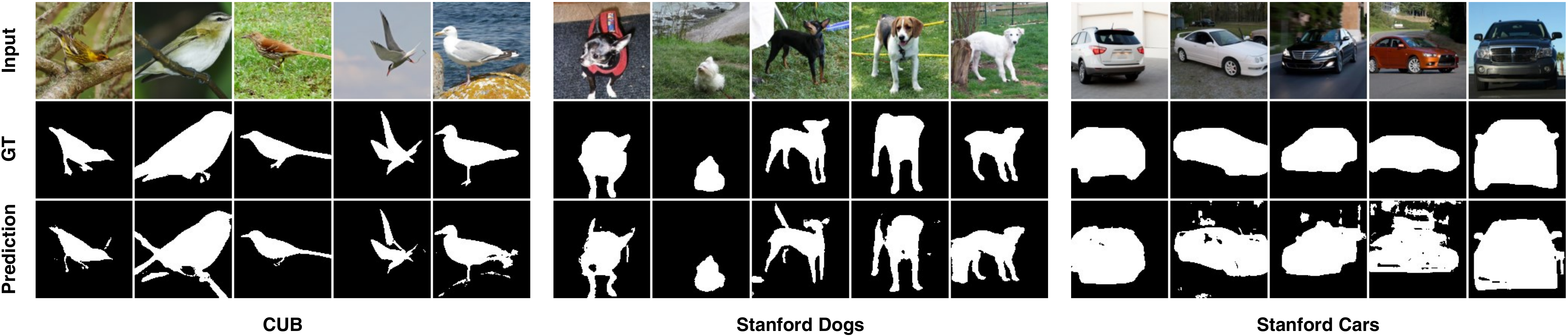}
	\caption{\textbf{Qualitative segmentation results} in CUB. The test images (Input), ground truth masks (GT), and predicted masks by segmentation network (Prediction), and  are presented.}
	\label{fig:qualitative}
\end{figure*}

\begin{table}[t]\centering
	\tablestyle{10pt}{1.2}\begin{tabular}{l|ccc}
		\shline
		\multirow{2}{*}{Methods} & \multicolumn{3}{c}{CUB} \\ 
		& ACC & IoU & mIoU \\
		\hline
		Supervised U-Net
		& 98.0 & 88.8 &	93.2 \\ 
		Voynov \etal~\cite{voynov2020big} 
		& 94.0 & 71.0 & - \\ 
		\hline
		GrabCut~\cite{rother2004grabcut} by~\cite{savarese2020information}  
		& 72.3 & 36.0 & 52.3 \\ 
		ReDO \cite{chen2019unsupervised} 
		& 84.5 & 42.6 & - \\ 
		PerturbGAN \cite{bielski2019emergence} 
		& - & - & 38.0 \\ 
		IEM~+~SegNet~\cite{savarese2020information}  
		& 89.3 & 55.1 & 71.4 \\ 
		Melas-Kyriazi~\etal~\cite{melas2021finding}
		& 92.1 & 66.4 & - \\ 
		\hline
		Ours  
		& \textbf{94.3} & \textbf{69.7} & \textbf{81.7} \\ 
		\shline
	\end{tabular}
	\caption{\textbf{Comparison to other unsupervised segmentation methods} in CUB. ``-'' indicates the unreported in original paper.}
	\label{tab:seg_benchmark}
\end{table}

Our methods are compared to other unsupervised segmentation methods on CUB at $128\times128$ resolution. 
Quantitative and qualitative results are presented in Table~\ref{tab:seg_benchmark} and \figref{fig:qualitative}, respectively.
It is noteworthy that the method proposed by Voynov~\etal~\cite{voynov2020big} requires manual inspection which should be regarded as external supervision. 
Our method outperform PerturbGAN~\cite{bielski2019emergence} with a large margin, showing the merit of our proposed mutual information maximization as discussed in the ablation study.
It is notable that performance level of our method is comparable to the state-of-the-art method~\cite{savarese2020information, melas2021finding}.
In particular, our method gain significantly higher performance than Melas-Kyriazi~\etal~\cite{melas2021finding} with 3.3 IoU on CUB. 

\subsection{Comparison to VAE-based methods}

\begin{table}[t]\centering
	\tablestyle{4.5pt}{1.25}\begin{tabular}{ll|ccc}
		\shline
		& Methods 		& FID~$\downarrow$ & ARI~$\uparrow$ & MSC~$\uparrow$ \\
		\hline
		\multirow{4}{*}{VAE-based} 
		& MONET-G~\cite{burgess2019monet}    & 269.3 & 0.11 & 0.48 \\
		& GENESIS~\cite{engelcke2019genesis} & 183.2 & 0.04 & 0.29 \\
		& SLOT-ATT.~\cite{locatello2020object} 	& - 	& 0.03 & 0.25 \\
		& GENESIS-V2~\cite{engelcke2021genesis} & 245.6 & 0.55 & 0.67 \\
		\hline
		GAN-based
		& Ours 									& \textbf{103.9} & \textbf{0.59} & \textbf{0.71} \\
		\shline
	\end{tabular}
	\caption{\textbf{Comparison to VAE-based methods} in APC.}
	\label{tab:apc_results}
\end{table}

\begin{figure}[t]\centering
	\includegraphics[width=0.95\linewidth]{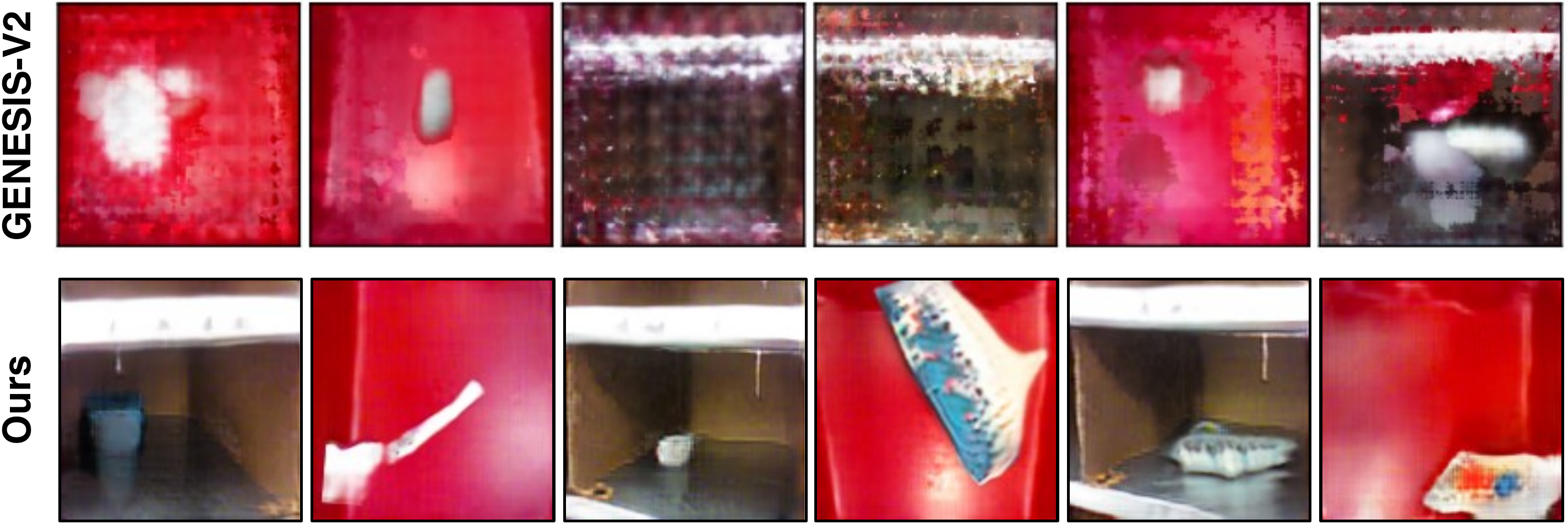}
	\caption{\textbf{Qualitative generation results} in APC. Our method can generate higher quality samples than GENESIS-V2, a VAE-based model.}
	\label{fig:apc_gen}
\end{figure}

\begin{figure}[t]\centering
	\includegraphics[width=0.95\linewidth]{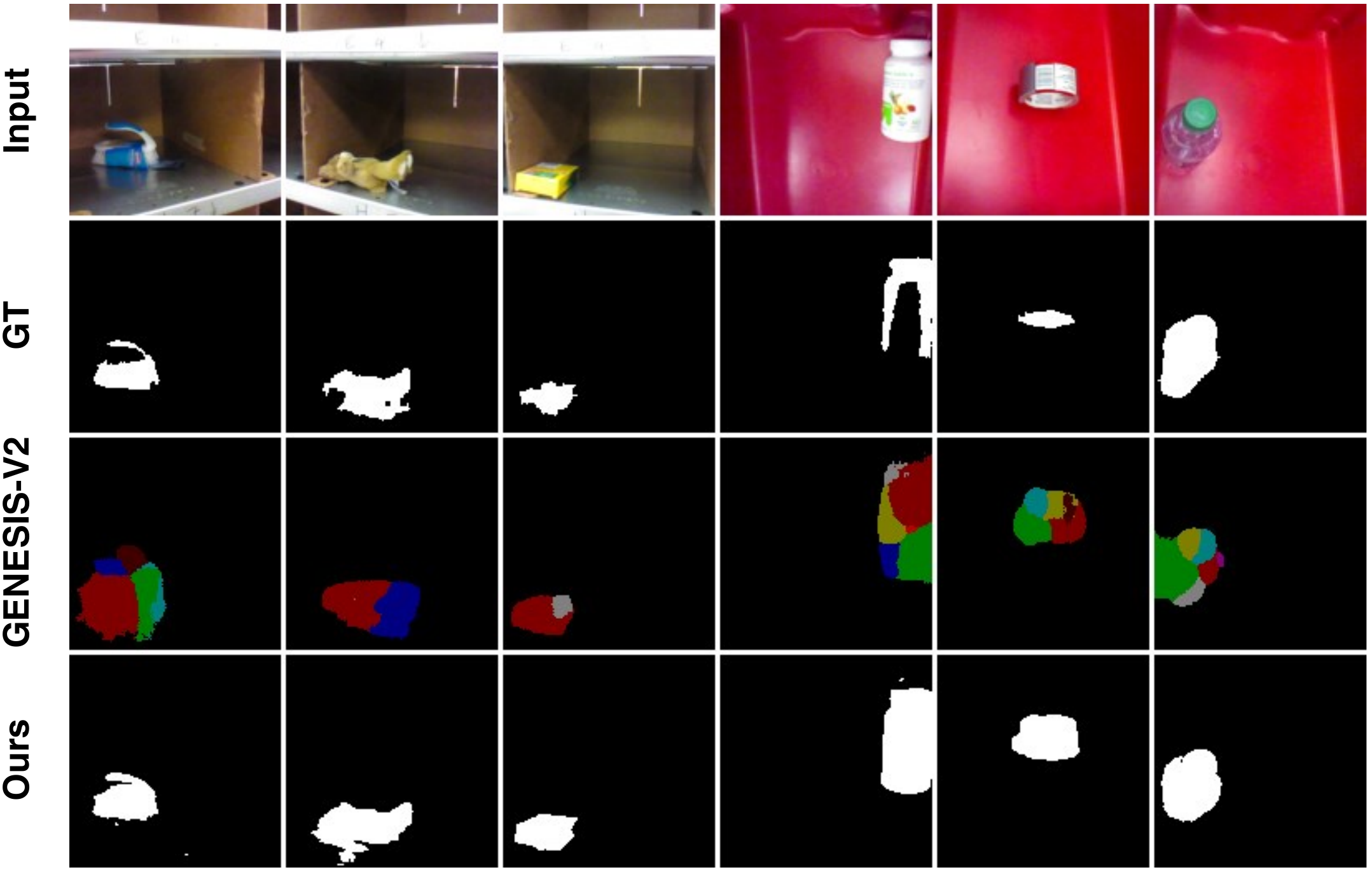}
	\caption{\textbf{Qualitative segmentation results} in APC. The test images (Input), ground truth masks (GT), predicted masks by GENESIS-V2 and our methods are presented.}
	\label{fig:apc_seg}
\end{figure}

Our methods are evaluated and compared to VAE-based models including MONet~\cite{burgess2019monet}, GENESIS~\cite{engelcke2019genesis}, SLOT ATTENTION~\cite{locatello2020object}, and GENESIS-V2~\cite{engelcke2021genesis} in APC dataset.
The quantitative and qualitative results are presented in Table~\ref{tab:apc_results}, \figref{fig:apc_gen}, and \figref{fig:apc_seg}.
The majority of VAE-based models perform poorly in realistic dataset as suggested in Table~\ref{tab:apc_results}.
GENESIS-V2 is the latest work that is successfully demonstrated in real-world dataset like APC. Therefore, our methods are mainly compared to GENESIS-V2. To obtain the results of GENESIS-V2, we use its released codes and models\footnote{\url{https://github.com/applied-ai-lab/genesis}}.

It can be seen from \figref{fig:apc_gen} and \figref{fig:apc_seg} that our methods are successfully applied to APC dataset which contains multi-category objects. These results show that our methods are not limited to single-category dataset. 
It is not surprising that our methods, as based on GAN, generate synthetic images of significant higher quality than GENESIS-V2 does, as suggested by the more visual pleasing images in \figref{fig:apc_gen} and much lower FID score in Table~\ref{tab:apc_results}.
Furthermore, our methods achieve higher segmentation performance than GENESIS-V2 and predict finer segmentation results than GENESIS-V2 does, as shown in \figref{fig:apc_seg}.

\subsection{Limitations}
Our generative model is limited to synthesize single object and single instance images. 
Our method also inherits the characteristic problems of GAN training including instability and requiring certain number of unlabelled training images.
In the future, we plan to extend our model into multi-instance image generation by modelling inter-object relationship and also improve the stability of training process with powerful architectures and techniques.

\section{Conclusion}
\label{sec:conclusion}
In this paper, we propose a method to improve the capability of layered GANs to disentangle foreground and background.
This improved layered GAN can successfully generate synthetic datasets to train segmentation networks which can be, in return, employed to further stabilize the training of layered GANs. 
Our method achieves competitive generation quality and segmentation performance compared to related methods on a variety of single-object image datasets.
We plan to extend our method to handle multiple object instances and the noisy data in future work.

\paragraph{Acknowledgement}
\label{sec:acknowledgement}
This work was supported by the National Key R\&D Program of China under Grant 2018AAA0102801, National Natural Science Foundation of China under Grant 61620106005, and EPSRC Visual AI grant EP/T028572/1.

{\small
\bibliographystyle{ieee_fullname}
\bibliography{egbib}

\begin{thebibliography}{10}\itemsep=-1pt

\bibitem{abdal2021labels4free}
Rameen Abdal, Peihao Zhu, Niloy Mitra, and Peter Wonka.
\newblock Labels4free: Unsupervised segmentation using stylegan.
\newblock {\em arXiv preprint arXiv:2103.14968}, 2021.

\bibitem{alexe2010classcut}
Bogdan Alexe, Thomas Deselaers, and Vittorio Ferrari.
\newblock Classcut for unsupervised class segmentation.
\newblock In {\em ECCV}, 2010.

\bibitem{anciukevicius2020object}
Titas Anciukevicius, Christoph~H Lampert, and Paul Henderson.
\newblock Object-centric image generation with factored depths, locations, and
  appearances.
\newblock {\em arXiv preprint arXiv:2004.00642}, 2020.

\bibitem{arandjelovic2019object}
Relja Arandjelovi{\'c} and Andrew Zisserman.
\newblock Object discovery with a copy-pasting gan.
\newblock {\em arXiv preprint arXiv:1905.11369}, 2019.

\bibitem{benny2020onegan}
Yaniv Benny and Lior Wolf.
\newblock Onegan: Simultaneous unsupervised learning of conditional image
  generation, foreground segmentation, and fine-grained clustering.
\newblock In {\em ECCV}, 2020.

\bibitem{bielski2019emergence}
Adam Bielski and Paolo Favaro.
\newblock Emergence of object segmentation in perturbed generative models.
\newblock In {\em NeurIPS}, 2019.

\bibitem{brock2018large}
Andrew Brock, Jeff Donahue, and Karen Simonyan.
\newblock Large scale gan training for high fidelity natural image synthesis.
\newblock In {\em ICLR}, 2019.

\bibitem{burgess2019monet}
Christopher~P Burgess, Loic Matthey, Nicholas Watters, Rishabh Kabra, Irina
  Higgins, Matt Botvinick, and Alexander Lerchner.
\newblock Monet: Unsupervised scene decomposition and representation.
\newblock {\em arXiv preprint arXiv:1901.11390}, 2019.

\bibitem{chai2011bicos}
Yuning Chai, Victor Lempitsky, and Andrew Zisserman.
\newblock Bicos: A bi-level co-segmentation method for image classification.
\newblock In {\em ICCV}, 2011.

\bibitem{chen2019unsupervised}
Micka{\"e}l Chen, Thierry Arti{\`e}res, and Ludovic Denoyer.
\newblock Unsupervised object segmentation by redrawing.
\newblock In {\em NeurIPS}, 2019.

\bibitem{chen2016infogan}
Xi Chen, Yan Duan, Rein Houthooft, John Schulman, Ilya Sutskever, and Pieter
  Abbeel.
\newblock Infogan: Interpretable representation learning by information
  maximizing generative adversarial nets.
\newblock In {\em NIPS}, 2016.

\bibitem{donahue2019large}
Jeff Donahue and Karen Simonyan.
\newblock Large scale adversarial representation learning.
\newblock In {\em NeurIPS}, 2019.

\bibitem{relate2020}
{S\'ebastien} Ehrhardt, Oliver Groth, Aron Monszpart, Martin Engelcke, Ingmar
  Posner, Niloy {J. Mitra}, and Andrea Vedaldi.
\newblock Relate: Physically plausible multi-object scene synthesis using
  structured latent spaces.
\newblock In {\em NeurIPS}, 2020.

\bibitem{engelcke2021genesis}
Martin Engelcke, Oiwi~Parker Jones, and Ingmar Posner.
\newblock Genesis-v2: Inferring unordered object representations without
  iterative refinement.
\newblock {\em arXiv preprint arXiv:2104.09958}, 2021.

\bibitem{engelcke2019genesis}
Martin Engelcke, Adam~R Kosiorek, Oiwi~Parker Jones, and Ingmar Posner.
\newblock Genesis: Generative scene inference and sampling with object-centric
  latent representations.
\newblock In {\em ICLR}, 2020.

\bibitem{eslami2016}
S.~M.~Ali Eslami, Nicolas Heess, Theophane Weber, Yuval Tassa, David
  Szepesvari, koray kavukcuoglu, and Geoffrey~E Hinton.
\newblock Attend, infer, repeat: Fast scene understanding with generative
  models.
\newblock In {\em NIPS}, 2016.

\bibitem{goetschalckx2019ganalyze}
Lore Goetschalckx, Alex Andonian, Aude Oliva, and Phillip Isola.
\newblock Ganalyze: Toward visual definitions of cognitive image properties.
\newblock In {\em ICCV}, 2019.

\bibitem{goodfellow2014generative}
Ian Goodfellow, Jean Pouget-Abadie, Mehdi Mirza, Bing Xu, David Warde-Farley,
  Sherjil Ozair, Aaron Courville, and Yoshua Bengio.
\newblock Generative adversarial nets.
\newblock In {\em NIPS}, 2014.

\bibitem{greff2019multi}
Klaus Greff, Rapha{\"e}l~Lopez Kaufmann, Rishabh Kabra, Nick Watters,
  Christopher Burgess, Daniel Zoran, Loic Matthey, Matthew Botvinick, and
  Alexander Lerchner.
\newblock Multi-object representation learning with iterative variational
  inference.
\newblock In {\em ICML}, 2019.

\bibitem{heusel2017gans}
Martin Heusel, Hubert Ramsauer, Thomas Unterthiner, Bernhard Nessler, and Sepp
  Hochreiter.
\newblock Gans trained by a two time-scale update rule converge to a local nash
  equilibrium.
\newblock In {\em NeurIPS}, 2017.

\bibitem{hochbaum2009efficient}
Dorit~S Hochbaum and Vikas Singh.
\newblock An efficient algorithm for co-segmentation.
\newblock In {\em ICCV}, 2009.

\bibitem{harkonen2020ganspace}
Erik Härkönen, Aaron Hertzmann, Jaakko Lehtinen, and Sylvain Paris.
\newblock Ganspace: Discovering interpretable gan controls.
\newblock In {\em NeurIPS}, 2020.

\bibitem{jahanian2019steerability}
Ali Jahanian, Lucy Chai, and Phillip Isola.
\newblock On the" steerability" of generative adversarial networks.
\newblock In {\em ICLR}, 2020.

\bibitem{ji2019invariant}
Xu Ji, Jo{\~a}o~F Henriques, and Andrea Vedaldi.
\newblock Invariant information clustering for unsupervised image
  classification and segmentation.
\newblock In {\em ICCV}, 2019.

\bibitem{joulin2010discriminative}
Armand Joulin, Francis Bach, and Jean Ponce.
\newblock Discriminative clustering for image co-segmentation.
\newblock In {\em CVPR}, 2010.

\bibitem{karras2019style}
Tero Karras, Samuli Laine, and Timo Aila.
\newblock A style-based generator architecture for generative adversarial
  networks.
\newblock In {\em CVPR}, 2019.

\bibitem{karras2020analyzing}
Tero Karras, Samuli Laine, Miika Aittala, Janne Hellsten, Jaakko Lehtinen, and
  Timo Aila.
\newblock Analyzing and improving the image quality of stylegan.
\newblock In {\em CVPR}, 2020.

\bibitem{khosla2011novel}
Aditya Khosla, Nityananda Jayadevaprakash, Bangpeng Yao, and Fei-Fei Li.
\newblock Novel dataset for fine-grained image categorization: Stanford dogs.
\newblock In {\em CVPR Workshop}, 2011.

\bibitem{kim2020unsupervised}
Wonjik Kim, Asako Kanezaki, and Masayuki Tanaka.
\newblock Unsupervised learning of image segmentation based on differentiable
  feature clustering.
\newblock {\em TIP}, 2020.

\bibitem{kingma2014adam}
Diederik~P Kingma and Jimmy Ba.
\newblock Adam: A method for stochastic optimization.
\newblock {\em arXiv preprint arXiv:1412.6980}, 2014.

\bibitem{kingma2014auto}
Diederik~P Kingma and Max Welling.
\newblock Auto-encoding variational bayes.
\newblock In {\em ICLR}, 2014.

\bibitem{krause20133d}
Jonathan Krause, Michael Stark, Jia Deng, and Li Fei-Fei.
\newblock 3d object representations for fine-grained categorization.
\newblock In {\em ICCV Workshop}, 2013.

\bibitem{liao2020towards}
Yiyi Liao, Katja Schwarz, Lars Mescheder, and Andreas Geiger.
\newblock Towards unsupervised learning of generative models for 3d
  controllable image synthesis.
\newblock In {\em CVPR}, 2020.

\bibitem{lin2020infogan}
Zinan Lin, Kiran Thekumparampil, Giulia Fanti, and Sewoong Oh.
\newblock Infogan-cr and modelcentrality: Self-supervised model training and
  selection for disentangling gans.
\newblock In {\em ICML}, 2020.

\bibitem{lin2020space}
Zhixuan Lin, Yi-Fu Wu, Skand~Vishwanath Peri, Weihao Sun, Gautam Singh, Fei
  Deng, Jindong Jiang, and Sungjin Ahn.
\newblock Space: Unsupervised object-oriented scene representation via spatial
  attention and decomposition.
\newblock In {\em ICLR}, 2020.

\bibitem{locatello2020object}
Francesco Locatello, Dirk Weissenborn, Thomas Unterthiner, Aravindh Mahendran,
  Georg Heigold, Jakob Uszkoreit, Alexey Dosovitskiy, and Thomas Kipf.
\newblock Object-centric learning with slot attention.
\newblock In {\em NeurIPS}, 2020.

\bibitem{melas2021finding}
Luke Melas-Kyriazi, Christian Rupprecht, Iro Laina, and Andrea Vedaldi.
\newblock Finding an unsupervised image segmenter in each of your deep
  generative models.
\newblock {\em arXiv preprint arXiv:2105.08127}, 2021.

\bibitem{miyato2018spectral}
Takeru Miyato, Toshiki Kataoka, Masanori Koyama, and Yuichi Yoshida.
\newblock Spectral normalization for generative adversarial networks.
\newblock In {\em ICLR}, 2018.

\bibitem{BlockGAN2020}
Thu Nguyen-Phuoc, Christian Richardt, Long Mai, Yong-Liang Yang, and Niloy
  Mitra.
\newblock Blockgan: Learning 3d object-aware scene representations from
  unlabelled images.
\newblock In {\em NeurIPS}, 2020.

\bibitem{niemeyer2021campari}
Michael Niemeyer and Andreas Geiger.
\newblock Campari: Camera-aware decomposed generative neural radiance fields.
\newblock {\em arXiv preprint arXiv:2103.17269}, 2021.

\bibitem{niemeyer2021giraffe}
Michael Niemeyer and Andreas Geiger.
\newblock Giraffe: Representing scenes as compositional generative neural
  feature fields.
\newblock In {\em CVPR}, 2021.

\bibitem{ouali2020autoregressive}
Yassine Ouali, C{\'e}line Hudelot, and Myriam Tami.
\newblock Autoregressive unsupervised image segmentation.
\newblock In {\em ECCV}, 2020.

\bibitem{plumerault2019controlling}
Antoine Plumerault, Herv{\'e} Le~Borgne, and C{\'e}line Hudelot.
\newblock Controlling generative models with continuous factors of variations.
\newblock In {\em ICLR}, 2020.

\bibitem{remez2018learning}
Tal Remez, Jonathan Huang, and Matthew Brown.
\newblock Learning to segment via cut-and-paste.
\newblock In {\em ECCV}, 2018.

\bibitem{ronneberger2015u}
Olaf Ronneberger, Philipp Fischer, and Thomas Brox.
\newblock U-net: Convolutional networks for biomedical image segmentation.
\newblock In {\em MICCAI}, 2015.

\bibitem{rother2004grabcut}
Carsten Rother, Vladimir Kolmogorov, and Andrew Blake.
\newblock " grabcut" interactive foreground extraction using iterated graph
  cuts.
\newblock {\em TOG}, 2004.

\bibitem{rother2006cosegmentation}
Carsten Rother, Tom Minka, Andrew Blake, and Vladimir Kolmogorov.
\newblock Cosegmentation of image pairs by histogram matching-incorporating a
  global constraint into mrfs.
\newblock In {\em CVPR}, 2006.

\bibitem{savarese2020information}
Pedro Savarese, Sunnie~SY Kim, Michael Maire, Greg Shakhnarovich, and David
  McAllester.
\newblock Information-theoretic segmentation by inpainting error maximization.
\newblock {\em arXiv preprint arXiv:2012.07287}, 2020.

\bibitem{saxe2013exact}
Andrew~M Saxe, James~L McClelland, and Surya Ganguli.
\newblock Exact solutions to the nonlinear dynamics of learning in deep linear
  neural networks.
\newblock {\em arXiv preprint arXiv:1312.6120}, 2013.

\bibitem{shen2020interpreting}
Yujun Shen, Jinjin Gu, Xiaoou Tang, and Bolei Zhou.
\newblock Interpreting the latent space of gans for semantic face editing.
\newblock In {\em CVPR}, 2020.

\bibitem{singh2019finegan}
Krishna~Kumar Singh, Utkarsh Ojha, and Yong~Jae Lee.
\newblock Finegan: Unsupervised hierarchical disentanglement for fine-grained
  object generation and discovery.
\newblock In {\em CVPR}, 2019.

\bibitem{spingarn2020gan}
Nurit Spingarn-Eliezer, Ron Banner, and Tomer Michaeli.
\newblock Gan steerability without optimization.
\newblock {\em ICLR}, 2021.

\bibitem{voynov2020unsupervised}
Andrey Voynov and Artem Babenko.
\newblock Unsupervised discovery of interpretable directions in the gan latent
  space.
\newblock In {\em ICML}, 2020.

\bibitem{voynov2020big}
Andrey Voynov, Stanislav Morozov, and Artem Babenko.
\newblock Big gans are watching you: Towards unsupervised object segmentation
  with off-the-shelf generative models.
\newblock {\em arXiv preprint arXiv:2006.04988}, 2020.

\bibitem{WahCUB_200_2011}
C. Wah, S. Branson, P. Welinder, P. Perona, and S. Belongie.
\newblock {The Caltech-UCSD Birds-200-2011 Dataset}.
\newblock Technical Report CNS-TR-2011-001, California Institute of Technology,
  2011.

\bibitem{yang2017lr}
Jianwei Yang, Anitha Kannan, Dhruv Batra, and Devi Parikh.
\newblock Lr-gan: Layered recursive generative adversarial networks for image
  generation.
\newblock {\em arXiv preprint arXiv:1703.01560}, 2017.

\bibitem{zeng2016multi}
Andy Zeng, Kuan-Ting Yu, Shuran Song, Daniel Suo, Ed Walker~Jr, Alberto
  Rodriguez, and Jianxiong Xiao.
\newblock Multi-view self-supervised deep learning for 6d pose estimation in
  the amazon picking challenge.
\newblock In {\em ICRA}, 2017.

\bibitem{zhang2020image}
Yuxuan Zhang, Wenzheng Chen, Huan Ling, Jun Gao, Yinan Zhang, Antonio Torralba,
  and Sanja Fidler.
\newblock Image gans meet differentiable rendering for inverse graphics and
  interpretable 3d neural rendering.
\newblock In {\em ICLR}, 2020.

\bibitem{zhang2021datasetgan}
Yuxuan Zhang, Huan Ling, Jun Gao, Kangxue Yin, Jean-Francois Lafleche, Adela
  Barriuso, Antonio Torralba, and Sanja Fidler.
\newblock Datasetgan: Efficient labeled data factory with minimal human effort.
\newblock In {\em CVPR}, 2021.

\end{thebibliography}
}

\clearpage

\renewcommand\thefigure{\thesection.\arabic{figure}}
\renewcommand\thetable{\thesection.\arabic{table}}
\setcounter{figure}{0}
\setcounter{table}{0}
\appendix
\onecolumn
\section{Network Details}\label{sec:impl_details}

\begin{figure}[t]
	\centering
	\includegraphics[width=\linewidth]{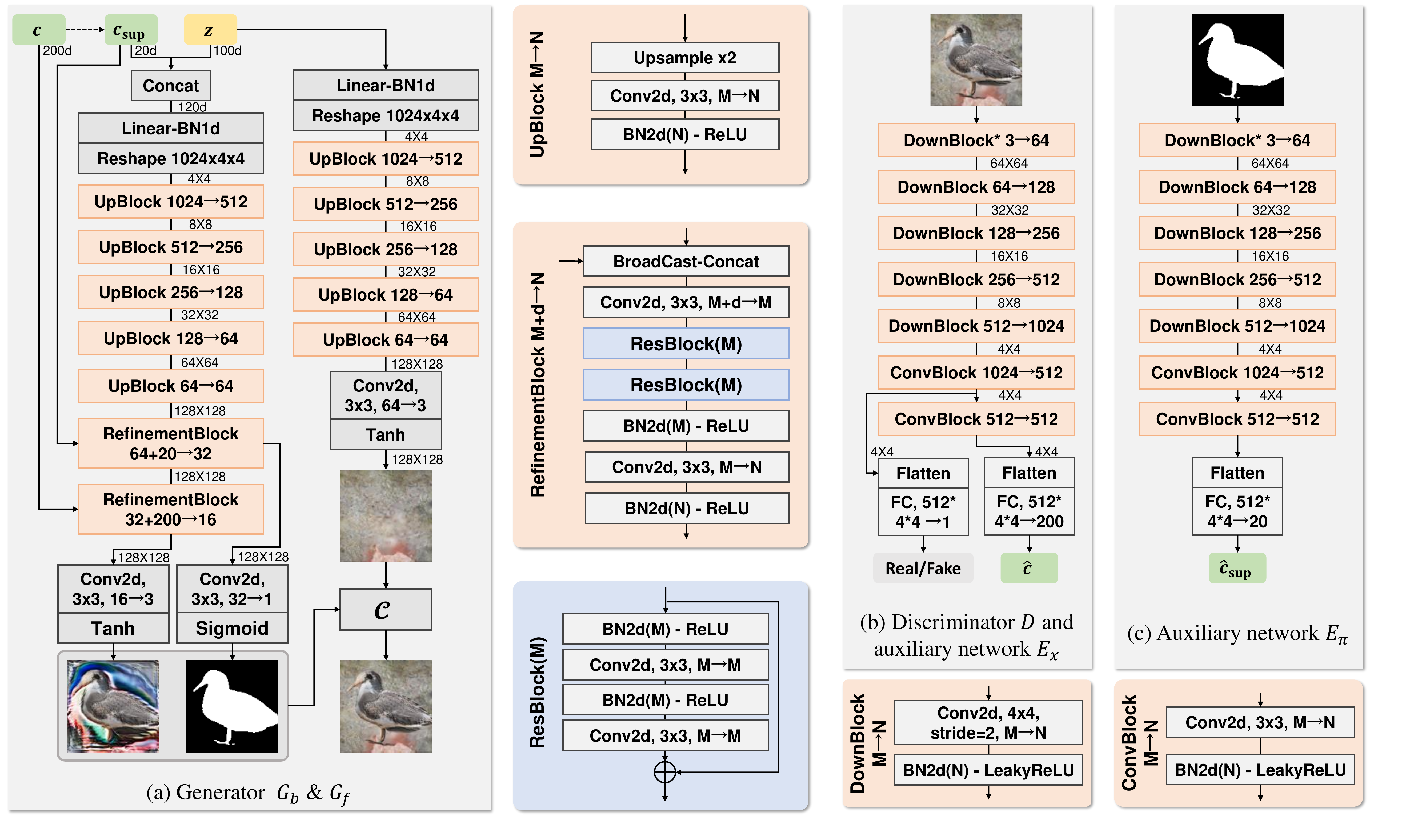}
	\caption{\textbf{Network structures}. Discriminator $D$ and the auxiliary network $E_x$ shares the backbone parameters and bifurcates at the top layers. ``DownBlock*'' denotes a DownBlock without BN layer.}
	\label{fig:structure}
\end{figure}

\paragraph{Latent codes} 
The public code $\rvz$ is continuous and randomly drawn from a normal distribution $N(0, 1)^{d_z}$. The private code $\rvc$ is designed as two-level codes, which is also known as parent and child code in FineGAN~\cite{singh2019finegan} and OneGAN~\cite{benny2020onegan}. In particular, $\rvc$ is discrete and randomly drawn from a categorical distribution $\text{Cat}(d_c)$, where $d_c$ denotes the number of categories. These categories are grouped into super categories so that every $\rvc$ can be associated with a super private code $\rvc_\text{sup}$. We use a very simple grouping mechanism that group categories with consecutive indices and the group size is varied dependent on datasets as in FineGAN~\cite{singh2019finegan} and OneGAN~\cite{benny2020onegan}.

\paragraph{Generator}
The structure of our generator (\figref{fig:structure}(a)) is adapted from FineGAN~\cite{singh2019finegan} which is tailored for generating images of fine-grained categories.
Compared to the original structure, the GRU activation function is replaced with ReLU to save some computation without sacrificing too much performance. 
It is noteworthy that only $\rvc_\text{sup}$ and $\rvz$ influences the generation process of foreground masks, while $\rvc_\text{sup}$, $\rvc$, and $\rvz$ have impacts on the foreground appearance, \ie RGB value.
The weights of all the convolutional and linear layers are initialized with orthogonal matrix~\cite{saxe2013exact}.

\paragraph{Perturbation} 
The perturbation $\mathcal{T}$ operates on any given image or mask, which is decomposed into three consecutive elementary transformation, isotropic scaling, rotation, and translation. Formally,
\begin{equation}\label{eq:affine_perturbation}
	\mathcal{T}(\rvx)(u, v, 1) = \rvx (\mA^{-1} (u, v, 1)),
	\mA = \mT(t_x, t_y)~\mR(\alpha)~\mS(s),~
\end{equation}
\begin{equation}
	\text{where}~
	\mS(s) = \left[{\arraycolsep=4pt\begin{array}{ccc}
			2^s & 0 & 0\\
			0 & 2^s & 0\\
			0 & 0 & 1
	\end{array}}\right], 
	\mT(t_x, t_y) = \left[{\arraycolsep=4pt\begin{array}{ccc}
			1 & 0 & t_x\\
			0 & 1 & t_y\\
			0 & 0 & 1
	\end{array}} \right],
	\mR(\alpha) = \left[{\arraycolsep=4pt\begin{array}{ccc}
			\cos\alpha & - \sin\alpha & 0\\
			\sin\alpha &   \cos\alpha & 0\\
			0 		   & 0 			  & 1
	\end{array}} \right],
\end{equation}
and the parameters of perturbation, $s$, $t_x$, $t_y$, and $\alpha$, are uniformly drawn from a small range. Besides the above geometric perturbation, we also use background contrast jittering as in PerturbGAN~\cite{bielski2019emergence}.

\paragraph{Discriminator}
The structure of our discriminator is presented in \figref{fig:structure}(b).
Spectral normalization~\cite{miyato2018spectral} is employed to stabilize the training.
The weights of all the convolutional and linear layers are initialized with orthogonal matrix~\cite{saxe2013exact}.

\paragraph{Auxiliary networks for mutual information maximization} 
\figref{fig:structure}(b)(c) presents the structures of our auxiliary networks for mutual information maximization.
Following previous work~\cite{chen2016infogan, lin2020infogan, singh2019finegan, benny2020onegan}, $E_x$ shares backbone with $D$ and bifurcates at the top layers.
$E_x$ takes as input an image and output a categorical distribution which is considered as a approximation of the posterior distribution $p(\rvc|\rvx)$.
$E_\pi$ has a similar role to $E_x$ but it approximates $p(\rvc_\text{sup}|\rvpi)$ as  $\rvc_\text{sup}$ controls the generation of masks.

\section{Dataset Details}
\paragraph{Caltech-UCSD Birds 200-2011 (CUB)} This dataset contains 11,788 bird images of 200 categories. It is split into 10,000 images for training, 788 images for validation, and 1,000 images for testing, as in~\cite{chen2019unsupervised}. All of the images only contain single instance and the interested objects are rarely occluded. The ground truth segmentation masks of the foreground object are annotated by humans and provided in the official release.

\paragraph{Stanford Dogs} This dataset contains 20,580 dog images of 120 categories. It is split into 12,000 images for training, and 8,580 images for testing by official release~\cite{khosla2011novel}. Most images contain single instance and a few images contain occluded objects and irrelevant objects like human may appear in the image. The ground truth segmentation masks are approximated with a Mask R-CNN pretrained on MSCOCO\footnote{We use the detectron2 library (\url{https://github.com/facebookresearch/detectron2}) and the model R101-FPN (\url{https://dl.fbaipublicfiles.com/detectron2/COCO-InstanceSegmentation/mask_rcnn_R_101_FPN_3x/138205316/model_final_a3ec72.pkl})}.

\paragraph{Stanford Cars} This dataset contains 16,185 car images of 196 categories. It is split into 8,144 images for training, and 8,041 images for testing by official release~\cite{krause20133d}. All if the images only contain single instance. The ground truth segmentation masks are approximated with a Mask R-CNN pretrained on MSCOCO, same model as used in Stanford Dogs.

\paragraph{Amazon Picking Challenge (APC)} This dataset\footnote{\url{https://vision.princeton.edu/projects/2016/apc/}} was created for evaluating 6D poses of objects in the warehouse environment. The authors also released an object segmentation training set which is used to pretrain the segmentation networks. We use the ``object segmentation training dataset''\footnote{\url{http://3dvision.princeton.edu/projects/2016/apc/downloads/training.zip}} for training our models. This dataset comprises images containing a single challenge object appearing either on a shelf or in a tray. 
Each object is shot as a series of scenes from different poses on both the shelf and in the red tray. Following GENESIS-V2~\cite{engelcke2021genesis}, we randomly select 10\% of scenes for validation, 10\% of scenes for testing and the rest for training using the script provided in GENESIS-V2 official source code\footnote{\url{https://github.com/applied-ai-lab/genesis}}.
The raw images are resized and center-cropped, resulting in $128\times128$ images.  The ground truth segmentation masks are provided in the official release. The ground truth masks are generated by certain automatic methods instead of annotated by humans. Therefore, these ground truth masks are quite noisy.

\section{Training Details}
\subsection{Training Layered GANs}
\paragraph{Data augmentation} The real images augmented with horizontal flip and random resized crop. This augmentation is used to increase the variation in training set and shown helpful to support perturbed composition~\cite{bielski2019emergence}. 

\paragraph{Optimization}
We employ Adam~\cite{kingma2014adam} optimizer with initial learning rate as $0.0002$ and $(\beta_1, \beta_2)$ as $(0.5, 0.999)$ train our layered GAN. The batch size is $32$. The training process lasts until 1,000,000 images are seen by discriminator.
Both the generator and the discriminator operate at $128\times128$ resolution.

\paragraph{Hyperparameters for each dataset} Table~\ref{tab:hyperparameter} lists the specific hyperparameters used for each dataset.

\begin{table}[h]\centering
\tablestyle{7pt}{1.25}\begin{tabular}{c|cccccccc}
\shline
				& $d_c$ & group size & scale $s$ & shift $t_x, t_y$ & rotation $\alpha$ & bg contrast & $\gamma_\text{mi}$ & $\gamma_\text{bg}$ \\
\hline	
CUB 			& 200 & 10 & $[-0.2, 0.]$ & $[-16\text{px}, 16\text{px}]$ 
				& $[-15^\circ, 15^\circ]$ & 0 & 1 & 2 \\
Stanford Dogs 	& 120 & 10 & $[-0.2, 0.]$ & $[-16\text{px}, 16\text{px}]$ 
				& $[-15^\circ, 15^\circ]$ & 0 & 2 & 1 \\
Stanford Cars	& 196 & 14 & $[-0.2, 0.]$ & $[-16\text{px}, 16\text{px}]$ 
				& $0^\circ$				  & $[0.7, 1.3]$ & 1 & 2 \\
APC 			& 200 & 10 & $[-0.2, 0.]$ & $[-16\text{px}, 16\text{px}]$ 
				& $[-15^\circ, 15^\circ]$ & 0 & 2 & 0 \\
\shline
\end{tabular}
\caption{\textbf{Hyperparameters} for training layered GANs on CUB, Stanford Dogs, Stanford Cars, and APC. }
\label{tab:hyperparameter}
\end{table}

\subsection{Alternate Training}
The training of layered GANs and the training of segmentation network is alternated for 5 rounds. In each round, the layered GANs are trained until discriminator has seen 2,000,000 images and the segmentation network is trained for 2,000 steps. Segmentation network is not involved into regularizing layered GANs until the second round. We employ a U-Net~\cite{ronneberger2015u} as segmentation network which is trained with Adam optimizer and the learning rate is 0.001 and the batch size is 32. Random color augmentation is used in training segmentation networks.

\subsection{Training Segmentation Networks}
A U-Net~\cite{ronneberger2015u} segmentation network is trained from the synthetic dataset with Adam optimizer. The initial learning rate is $0.001$ and batch size is $32$. The training duration is 12,000 steps. Random color augmentation is used. The During inference, following~\cite{chen2019unsupervised, savarese2020information}, the input image and ground truth masks are rescaled and center cropped to $128\times128$.

\section{More Results}

\paragraph{Ablation study \wrt mutual information maximization}

\begin{table}[h]\centering
	\tablestyle{4pt}{1.25}\begin{tabular}{c|ccccc}
		\shline
		\multicolumn{6}{c}{Stanford Cars} \\
		\hline
		& \multicolumn{5}{c}{$\gamma_\text{mi}$} \\
		& 0.001 & 0.01 & 0.1 & 0.5 & 1.0 \\
		\hline	
		FID~$\downarrow$ 	& 16.0 & 14.8 & 14.9 & \textbf{14.6} & 14.7 \\
		bg-FID~$\uparrow$ 	& 145.0 & 134.4 & 162.9 & 168.2 & \textbf{180.3} \\
		IoU~$\uparrow$ 		& 45.2 & 49.0 & 54.8 & \textbf{59.7}  & 58.7  \\
		\shline
	\end{tabular} ~~~~~~~
\tablestyle{4pt}{1.25}\begin{tabular}{c|ccccc}
	\shline
	\multicolumn{6}{c}{Stanford Dogs} \\
	\hline
	& \multicolumn{5}{c}{$\gamma_\text{mi}$} \\
	& 0.1 & 0.5 & 1.0 & 5.0 & 10.0 \\
	\hline	
	FID~$\downarrow$ 	& 47.3 & 36.8 & \textbf{34.5} & 46.5 & 52.1 \\
	bg-FID~$\uparrow$ 	& 46.9 & 106.6 & 113.3 & 115.6 & \textbf{117.5} \\
	IoU~$\uparrow$ 		& 36.8 & 58.0 & 62.1 & \textbf{64.8} & 57.1 \\ 
	\shline
\end{tabular}
	\caption{\textbf{Ablation study with respect to mutual information maximization} on Stanford Cars and Stanford Dogs at $64\times64$ resolution. $\gamma_\text{mi}$ denotes the loss weight of mutual information maximization. The alternate training is disabled.}
	\label{tab:app_ablation_mi}
\end{table}

We further present the ablation study with respect to mutual information maximization on Stanford Cars and Stanford Dogs in Table~\ref{tab:app_ablation_mi} at resolution $64\times64$.
It can be concluded that an appropriate $\gamma_\text{mi}$ is essential to achieve optimal performance, while a wide range of $\gamma_\text{mi}$ can assure the success of learning. Notably, the appropriate $\gamma_\text{mi}$ might be different from dataset to dataset. For example, on Stanford Cars, even when the $\gamma_\text{mi}$ is as low as $0.001$, the segmentation is still learned though with low performance (IoU $45.2$).
However, on Stanford Dogs and CUB, a very low $\gamma_\text{mi}$ (\eg $0.1$) would lead to the failure of disentangling foreground and background.

\end{document}